\documentclass{ieeeaccess}
\usepackage{cite}
\usepackage{amsmath,amssymb,amsfonts}
\usepackage{graphicx}
\usepackage{textcomp}
\usepackage{color,soul}

\usepackage{algorithm} 
\usepackage{algorithmic} 

\newcommand{\etal}{\textit{et al.}}
\newcommand{\eg}{\textit{e}.\textit{g}.}

\def\BibTeX{{\rm B\kern-.05em{\sc i\kern-.025em b}\kern-.08em
    T\kern-.1667em\lower.7ex\hbox{E}\kern-.125emX}}
\begin{document}
\history{Date of publication xxxx 00, 0000, date of current version xxxx 00, 0000.}
\doi{10.1109/ACCESS.2017.DOI}

\title{Human and Scene Motion Deblurring using Pseudo-blur Synthesizer}
\author{\uppercase{Jonathan Samuel Lumentut}\authorrefmark{1}, \IEEEmembership{Member, IEEE},
\uppercase{In Kyu Park}.\authorrefmark{1}, \IEEEmembership{Senior Member, IEEE}}
\address[1]{Department of Information and Communication Engineering, Inha University, Incheon 22212, Korea (e-mail: jlumentut@gmail.com, pik@inha.ac.kr)}
\tfootnote{This work was supported by Samsung Research Funding Center of Samsung Electronics under Project Number SRFCIT-1901-06. This work was supported by Institute of Information \& communications Technology Planning \& Evaluation~(IITP) grant funded by the Korea government~(MSIT)~(2020-0-01389, Artificial Intelligence Convergence Research Center~(Inha University)). This work was supported by Inha University Research Grant.}

\markboth
{J. S. Lumentut \headeretal: Human and Scene Motion Deblurring using Pseudo-blur Synthesizer}
{J. S. Lumentut \headeretal: Human and Scene Motion Deblurring using Pseudo-blur Synthesizer}

\corresp{Corresponding author: In Kyu Park (e-mail: pik@inha.ac.kr)}

\begin{abstract}
Present-day deep learning-based motion deblurring methods utilize the pair of synthetic blur and sharp data to regress any particular framework. This task is designed for directly translating a blurry image input into its restored version as output. The aforementioned approach relies heavily on the quality of the synthetic blurry data, which are only available before the training stage. Handling this issue by providing a large amount of data is expensive for common usage. We answer this challenge by providing an on-the-fly blurry data augmenter that can be run during training and test stages. To fully utilize it, we incorporate an unorthodox scheme of deblurring framework that employs the sequence of blur-deblur-reblur-deblur steps. The reblur step is assisted by a reblurring module (synthesizer) that provides the reblurred version (pseudo-blur) of its sharp or deblurred counterpart. The proposed module is also equipped with hand-crafted prior extracted using the state-of-the-art human body statistical model. This prior is employed to map human and non-human regions during adversarial learning to fully perceive the characteristics of human-articulated and scene motion blurs. By engaging this approach, our deblurring module becomes adaptive and achieves superior outcomes compared to recent state-of-the-art deblurring algorithms.
\end{abstract}


\titlepgskip=-15pt

\maketitle

\section{Introduction}
The idea of recovering blurry images into their sharp version has been presented since a decade ago and remains an active research area in computer vision. 
The spread of pixels usually causes a blurry image owing to the motion effect during capture time. 
This motion is modeled by a specific point spread function~(PSF) and can be represented as a blur kernel. 
Early motion-blurred image is modeled by the blur kernel that is directly convolved to a sharp image with few additive noises. 
The task of restoring PSF-based degradation is known as motion deblurring. 
Based on the previous assumption, traditional methods solve motion deblurring by deconvolving back the blurry image with a predicted blurry kernel. 
This idea was implemented in famous state-of-the-art works~\cite{Sunghyun_SIGGRAPH2009,Whyte_CVPR2010,Xu_ECCV2010,Krishnan_CVPR2011,Whyte_IJCV2012,Pan_CVPR2014,Pan_CVPR2016} where various regularization priors are also advocated to help the deblurring procedures.
With the rise of deep learning, many kernel-free deblurring works are introduced.
The work of Nah~\etal~\cite{Nah_CVPR2017} and Kupyn~\etal~\cite{Kupyn_CVPR2018, Kupyn_ICCV2019} utilize generative-adversarial-network (GAN)~\cite{Goodfellow_NIPS2014} to solve this issue.
The recent deep-learning based deblurrings are improved by adopting feature-level modification~\cite{ZhangStack_CVPR2019,Suin_CVPR2020,Shen_ICCV2019_Humanaware} and region-based prior utilization~\cite{shenface_CVPR18, Shen_ICCV2019_Humanaware, Lumentut_ACCV2020}.

Following the vast growth of image restoration works, we observe that the current highlight involves an unorthodox approach for improving performance.
Early work by Chen~\etal~\cite{ChenReb2Deb_ICCP2018} solves deblurring by reblurring the deblur output.
The reblur output is being supervised with the blurred input during training.
To simplify, Chen~\etal~\cite{ChenReb2Deb_ICCP2018} apply the sequence of \textit{blur-deblur-reblur} ($\textit{B}\rightarrow\textit{D}\rightarrow\textit{R}$) in the training scheme with $R \approx B$, while maintaining the ($\textit{B}\rightarrow\textit{D}$) in the test scheme. 
Recently, Zhang~\etal~\cite{ZhangRealist_CVPR2020} provide a unique approach by supplying a reblur network at the top of the deblurring module for unpaired strategy.
Their approach~\cite{ZhangRealist_CVPR2020} employs noise-based re-blurred version of any sharp image $R_{\mathcal{N}}$ and utilize them in the training process ($R_{\mathcal{N}}\rightarrow\textit{D}$).
Take note that $R_{\mathcal{N}}$ will be inconsistent as noise is generated randomly.
Thus, they~\cite{ZhangRealist_CVPR2020} also still perform ($\textit{B}\rightarrow\textit{D}$) sequence in the test scheme without utilizing $R_{\mathcal{N}}$.
These approaches indicate that re-corrupting initial information is useful for augmenting the data.

From this motivation, we introduce the order of \textit{blur-deblur-reblur-deblur} ($\textit{B}\rightarrow\textit{D}\rightarrow\textit{R}\rightarrow\textit{D}$) as a modish approach to solve human and scene motion deblurring. 
Unlike the previous unorthodox methods~\cite{ChenReb2Deb_ICCP2018,ZhangRealist_CVPR2020}, our sequence is performed in \textbf{training} and \textbf{test} stages.
This approach allows our deblurring method to learn the augmented blurry data $R$  that are \textbf{different} from the blurry input $B$ (with $R \not\approx B$) in \textbf{both} stages.
To obtain a consistent $R$, we provide a reblurring module (pseudo-blur synthesizer) that only receives a single RGB image, which is trained with the localized regions of the human body and scene.
This strategy is applied as human-articulated and scene motion display different blur characteristics.
Our motivation is that, up to recent time, only a few pioneer works~\cite{Shen_ICCV2019_Humanaware,Lumentut_ACCV2020} that particularly handle the deblurring on human body case.
The idea of human deblurring is presented by Shen~\etal~\cite{Shen_ICCV2019_Humanaware} that utilizes separated foreground and background maps to distinguish human and non-human regions.
However, this approach is un-precise as they produce pre-generated rectangular maps to cover the human body.
Moreover, in their case, the blur is unlikely to represent the motion blur caused by human articulated body joint movement.
This approach is tackled by Lumentut~\etal~\cite{Lumentut_ACCV2020} by producing a localized map that covers both \textit{human body} and \textit{its nearby regions} that are affected by the \textit{body-joint articulation motion}.
In this work, the idea of the localized prior map is utilized in the \textit{reblurring} procedure as part of the novel sequence.
To achieve it, we propose an adversarial-based framework that learns both scene and human motion blur characteristics for supplying the reblurring module.

Once the reblurring module is settled, $R$ is treated as augmentation data. 
However, the ultimate goal of adopting the proposed sequence is to perform self-adaptation in the test stage.
This is important as many restoration models are trained with a limited dataset but required to solve various issues.
To achieve it, the proposed sequence is plugged in a model-agnostic meta-learning algorithm~\cite{FinnMAML_2017}, that shows significant performance improvement in previous non-deblurring studies~\cite{KimFast_ECCV2020,ChoiScene_CVPR2020,SohMeta_CVPR2020}.
By implementing this strategy, we show that our approach is superior in deblurring the real-world scenario, where no related training data is available.
This benefit is obtained due to the presence of our pseudo-blur synthesizer that supports the proposed sequence.
To summarize, we describe our contributions as 3 manifolds:
\begin{itemize}
\item We present a unique sequence of deblurring in training and testing procedures, which allows a self-adapting capability that yields superior results compared to recent state-of-the-art works.
\item We provide, to our best knowledge, a novel way to synthesize a blurry image from only a single RGB image input, achieved by employing localized human and non-human regions of an image.
\item We show that the hand-crafted human-prior in the reblurring module is learnable via adversarial strategy, subsequently improving the deblurring performance.
\end{itemize}

\section{Related Works}
\paragraph{Motion deblurring}
Early deblurring algorithms utilize the classical way of restoration by firstly estimating the blur kernel.
The estimated kernel is used to deconvolve the blurry input to obtain sharp input. 
Various regularization priors are utilized for improving this approach~\cite{Sunghyun_SIGGRAPH2009,Whyte_CVPR2010,Xu_ECCV2010,Krishnan_CVPR2011,Whyte_IJCV2012,Pan_CVPR2014,Pan_CVPR2016,LiuRyan_ICASSP2018_l0reg}. 
These works further target multi-view imaging, as shown in these studies~\cite{Sellent_ECCV16,Srinivasan_CVPR2017,Lumentut_SPL2019}.
A recent trend on deep learning moves toward GAN-based~\cite{Goodfellow_NIPS2014} architecture for its capability to directly translate an image to a certain domain~\cite{Isolapix2pix_CVPR2017}.
This approach is then followed by these notable deblurring works~\cite{Nah_CVPR2017, Kupyn_CVPR2018,Kupyn_ICCV2019}. 
Recent works in deblurring include hand-crafted priors to prioritize certain regions for learning. 
These works are shown by Shen~\etal~\cite{shenface_CVPR18} and Ren~\etal~\cite{Ren_facedeb_3Dprior_ICCV2019} for face deblurring as well as Shen~\etal~\cite{Shen_ICCV2019_Humanaware} and Lumentut~\etal~\cite{Lumentut_ACCV2020} for human deblurring. 
As explained in the previous section, the recent works of non-classic restoration (non $\textit{B}\rightarrow\textit{D}$ in training sequence)~\cite{ChenReb2Deb_ICCP2018, ZhangRealist_CVPR2020} capture our attention. 
These approaches show that the addition of a reblurring step improves the deblurring performance.
Both closely related works of Shen~\etal~\cite{Shen_ICCV2019_Humanaware} and Lumentut~\etal~\cite{Lumentut_ACCV2020} inspire our works to solve human deblurring.
As described in the previous section, our method takes advantage of the prior map in generating realistic augmented blurry data via reblurring module.
We utilize this module to help the deblurring module in achieving its self-adaptive capability.

\paragraph{Synthetic blur generation}
The early work of generating a blurry image is initiated by~\cite{Kohler_ECCV2012}. 
Their work produces camera motion with a robot system that moves the camera in an accurate position. This approach is bulky and hard to be applied in daily use. 
A more complex blurry dataset is introduced by~\cite{Lai_CVPR2016} in which sets of motion blurs that are recorded using the inertial sensor of a consumer cellphone are collected. 
These motion blurs are convolved directly to the sharp image to produce a synthetic blurry image. Recent approaches show that the averaging multiple-frames is faithful enough to generate realistic scene blur as expressed by~\cite{Nah_CVPR2017}. 
Another non-typical approach is introduced by Brooks and Barron~\cite{brooks_cvpr2019} wherein a blurry image is produced from the two successive sharp frames. 
Their idea is based on frame interpolation work as it produced several intermediate frames within the two inputs to generate a smooth blurry result.
The recent work of~\cite{Shen_ICCV2019_Humanaware} provides blurry human images as their dataset to solve a particular issue, notably human deblurring.
However, their blur result is affected by non-human articulated motion. 
Unlike~\cite{Shen_ICCV2019_Humanaware}, we consider both constraints on generating the synthetic blur achieved by pursuing the local human body region.
The closest work to ours is by Zhang~\etal~\cite{Zhang_CVPR2019} that provides a reblurring network to blur the sharp image during training ($\textit{R}\rightarrow\textit{D}$).
This method, however, relies fully on the additional noise at the input.
On the contrary, our blur synthesizer network only requires a single red-green-blue (RGB) input, which guarantees a consistent reblur output.

\paragraph{Meta-learning application}
The meta-learning approach paves a unique way for recent restoration works~\cite{KimFast_ECCV2020,ChoiScene_CVPR2020,SohMeta_CVPR2020}.
Its objective is to provide an updated version of a network that is adaptive during test time.
In general, meta-learning is categorized into three groups. 
The first group belongs to the metric-based method~\cite{vinyal_nips2016, snellproto_nips2017}. 
This approach has the objective of seeking metric space that provides efficient learning under a few samples. 
The second group belongs to the memory network-based approach~\cite{santoro_icml2016,oreshkin_nips2016,mishra_iclr2018} where its objective is to train a network that learns across various tasks to be robust to the unseen task. 
The last group belongs to the optimization-based approach, wherein gradient-based learning is employed. 
The main idea is to find an initial transferable point that helps the network adapt within a few gradient updates~\cite{grant_iclr2018,FinnMAML_2017,finnother_iclr2018}. 
The recent model-agnostic meta-learning~\cite{FinnMAML_2017} method that utilizes gradient descent learning shows a significant impact on super-resolution studies~\cite{SohMeta_CVPR2020,KimFast_ECCV2020}. 
Just recently, a test-time adaptive version of the motion deblurring method is proposed by Chi~\etal~\cite{Chi_TestTimeDeblur_CVPR2021}.
This work is closely related to ours, however they only utilize ($\textit{B}\rightarrow\textit{D}\rightarrow\textit{R}$) sequence similar to~\cite{ChenReb2Deb_ICCP2018} which enforces $R$ to be equal to $B$. 
As opposed to this setup, we employ our reblurrer to synthesize $R$ differently to $B$ as its augmented version.
In our experiments, we show that our proposed strategy succeeds in enhancing the deblurring network via meta-learning.

\begin{figure*}[t]
\begin{center}
        \includegraphics[width=1.0\textwidth]{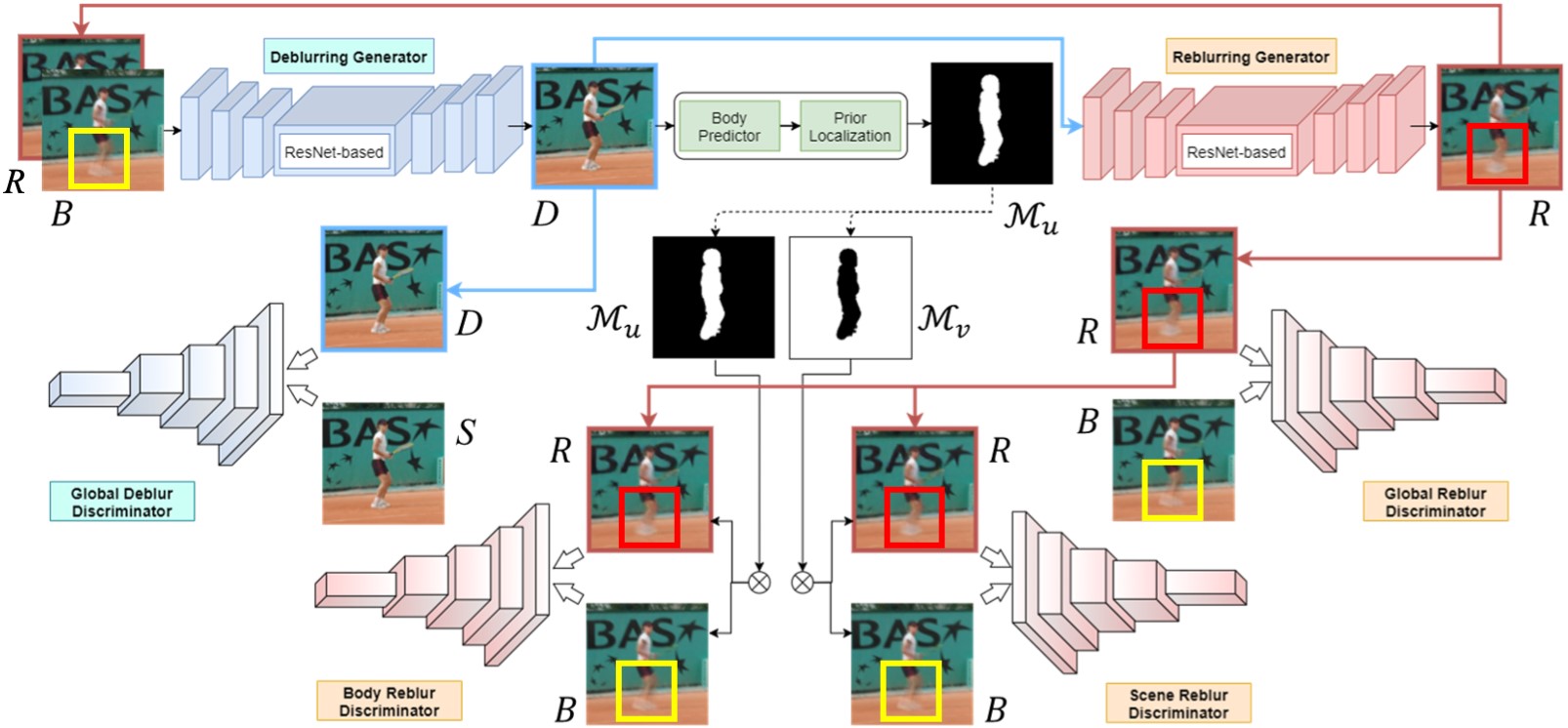}
\end{center}
\vspace{-0.2cm}
\caption{The figure above represents the framework of our deblurring-reblurring tasks. Blue modules are utilized for training the deblurring procedure, while red modules are used to train the reblurring procedure. As emphasized in the smaller rectangular boxes (red and yellow boxes), our synthesized blur result, $R$, is \textbf{different} with blurry input $B$ ($R \not\approx B$). Moreover, $R$ is deblurred again in our algorithm to induce the self-adaptive capability. Viewing it on an \textit{electronic screen} is advised.}
\label{fig:fig_mainframework}
\end{figure*}

\begin{figure*}[t]
\begin{center}
        \includegraphics[width=1.0\textwidth]{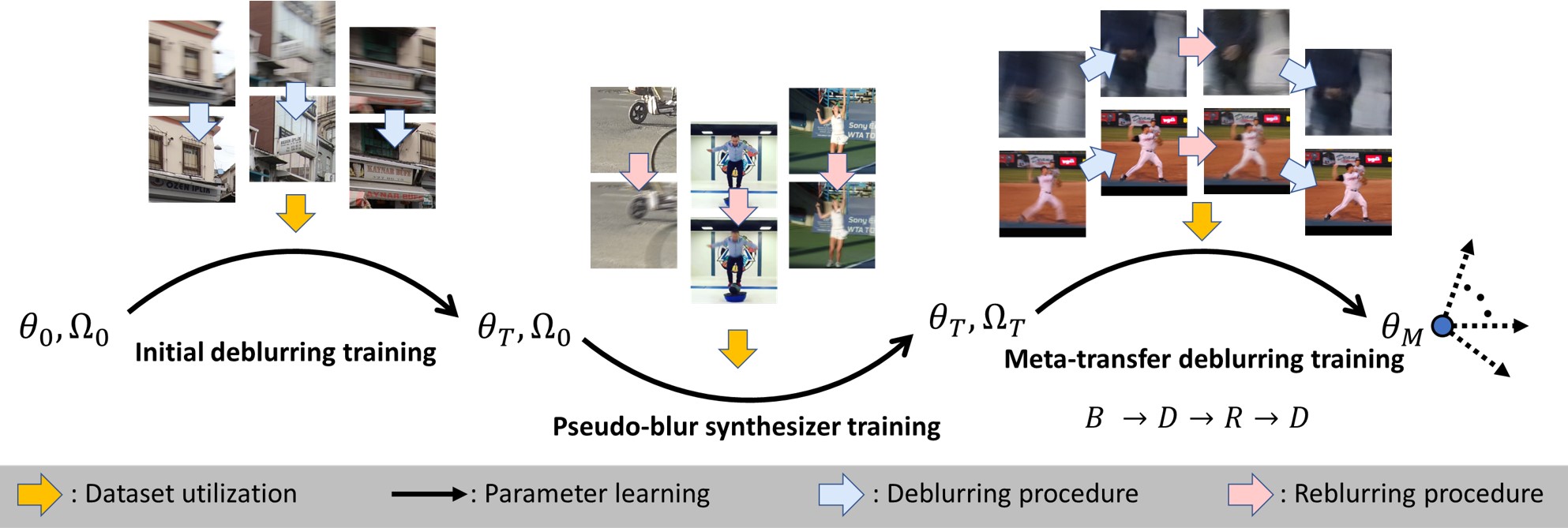}
\end{center}
\vspace{-0.2cm}
\caption{The main scheme of our \textbf{training} phase. The training procedures are run in a progressive stages, namely: \textit{initial deblurring training}, \textit{pseudo-blur synthesizer training}, and \textit{meta-transfer deblurring training}. In the initial stage, only deblurring modules (blue objects) are trained. Then, in the pseudo-blur synthesizer training stage, only the reblurring modules (red objects) are trained. In the meta-transfer deblurring training stage, the deblurring modules are furtherly optimized with the utilization of frozen reblurrer module. Note that, in the final stage, no discriminators are further optimized.}
\label{fig:fig_meta_training_scheme}
\end{figure*}

\section{Method}
We introduce a set of procedures for training (Figure~\ref{fig:fig_meta_training_scheme}) and testing (Figure~\ref{fig:fig_meta_testing_scheme}) stages that fully utilize the proposed sequence.
Initially, the deblurring network is trained until it converges ($\theta_T$).
The deblurring network is then frozen for training the reblurring network ($\Omega_T$).
These outcomes are utilized in our meta-transfer-learning procedure to obtain a ready-to-be-adapted deblurring weight ($\theta_M$).
Finally, the meta-testing procedure is employed to run the self-adaptation strategy ($\theta_k$).
In these subsections, we describe briefly on each particular scheme: \textit{initial deblurring training}, \textit{pseudo-blur synthesizer training}, \textit{meta-transfer learning}, and \textit{meta-testing} procedures.

\subsection{Initial Deblurring Training}
\label{lab:deblur_parag}
The whole process is started by initially train the deblurring network. 
This network is trained with the involvement of GoPro~\cite{Nah_CVPR2017} and HIDE~\cite{Shen_ICCV2019_Humanaware} dataset. 
As displayed in Figure~\ref{fig:fig_mainframework}, the deblurring network receives an input of blurry RGB image $B$ and produces deblurred output $D$.
In this stage, training the deblurring is collaborated with the \textit{global deblur} discriminator module.
The term $global$ represents the utilization of full image region.
As shown in Figure~\ref{fig:fig_mainframework}, the deblurring module is represented by the~\textit{deblurring generator} annotation.
~\textit{Global deblur discriminator} module is also utilized to influence the generator.
These networks, highlighted with blue color, are trained using GoPro~\cite{Nah_CVPR2017} and HIDE~\cite{Shen_ICCV2019_Humanaware} dataset.
The configuration of~\textit{deblurring generator}\footnote{The reblurred image $R$ is mentioned in Table~\ref{table:tab_detail_deblurring_generator} as our naive-finetuned version (Ours-\textbf{F}) utilizes the deblurring module to deblur $R$ in the sequence of $\textit{B}\rightarrow\textit{D}\rightarrow\textit{R}\rightarrow\textit{D}$ during training stage (see the Ablation Study in the subsection~\ref{sec:subsec_abla_study})} and~\textit{global deblur discriminator} networks are provided in Table~\ref{table:tab_detail_deblurring_generator} and Table~\ref{table:tab_detail_deblurring_discriminator}, respectively.
9 Residual Blocks~\cite{He_CVPR2016} configuration is stacked to convey intermediate features.

\noindent\textbf{Optimization of deblurring module}
To optimize the deblurring network in the \textit{initial deblurring training} stage, we utilize a simple absolute error calculation between deblurred output $D$ and sharp ground truth image $S$ within a mini-batch $b$, represented as $(L^{deb}_{1} = \frac{1}{b}\sum_{c}^{b}\| S_{c} - D_{c}\|)$, where $c$ represents data on each batch. 
This loss is countered with a single \textit{global deblur discriminator} as shown in Figure~\ref{fig:fig_mainframework}. 
The discriminator receives the input of both sharp $S$ for the real case and deblurred $D$ for the fake case following the recent least-square GAN (LSGAN) introduced in~\cite{Mao_ICCV2017_LSGAN}. 
In detail, the adversarial real and fake losses of the deblurring discriminator are represented as:
\begin{equation} 
L^{deb}_{\textbf{Real}} = \frac{1}{b} \sum_{c}^{b} \Pi_{\textbf{Glo}}(S_{c});
L^{deb}_{\textbf{Fake}} = \frac{1}{b} \sum_{c}^{b} \Pi_{\textbf{Glo}}(D_{c}),
\label{eq:eq_DebRealFake}
\end{equation}
where $\Pi(\cdot)$ represents the discriminative function. 
The two functions are combined in the generator and discriminator losses of the deblurring, which are written as:
\begin{equation} 
L^{deb}_{\textbf{Disc}} = 0.5 \times ( \| L^{deb}_{\textbf{Real}} - 1\|^2 + \| L^{deb}_{\textbf{Fake}}\|^2);
\label{eq:eq_ScoreDisc_Deb}
\end{equation}
\begin{equation} 
L^{deb}_{\textbf{Gen}} = L^{deb}_{1} + 0.5 \times (\| L^{deb}_{\textbf{Fake}} - 1\|^2).
\label{eq:eq_ScoreGen_Deb}
\end{equation}
Note that our discriminator requires an input image that is fully divided by 16; thus, we utilize a patch size of 128$\times$128 in the training procedures.
\begin{table}[t]
\huge
\begin{center}
\caption{Detailed settings of our deblurring generator module. The spatial output size is downsampled and upsampled according to the stride number.}
\vspace{-0.2cm}
\label{table:tab_detail_deblurring_generator}
\resizebox{\columnwidth}{!}{
\begin{tabular}{llcc}
\hline\noalign{\smallskip}
Layer $\qquad\qquad$& Detail & Output size & Stride\\
\hline\noalign{\smallskip}
Input ($B$ or $R$) $\qquad\qquad$ & - $\qquad$& ($H \times W \times 3$) & -\\
Conv \(7 \times 7\) & \textit{IN}+\textit{ReLU} $\qquad$& ($H \times W \times 64$) & 1\\
Conv \(3 \times 3\)& \textit{IN}+\textit{ReLU} $\qquad$& ($H/2 \times W/2 \times 64$) & 2\\
Conv \(3 \times 3\) & \textit{IN}+\textit{ReLU} $\qquad$& ($H/4 \times W/4 \times 128$) & 2\\
Res\_Blocks\_1-9 \(3 \times 3\) & \textit{IN}+\textit{ReLU} $\qquad$& ($H/4 \times W/4 \times 256$) & 1\\
ConvTrans \(3 \times 3\) & \textit{IN}+\textit{ReLU} $\qquad$& ($H/2 \times W/2 \times 128$) & 2\\
ConvTrans \(3 \times 3\) & \textit{IN}+\textit{ReLU} $\qquad$& ($H \times W \times 64$) & 2\\
Conv \(7 \times 7\) & \textit{Tanh} $\qquad$& ($H \times W \times 3$) & 1\\
\hline
\noalign{\smallskip}
\end{tabular}
}
\end{center}
\end{table}
\begin{table}[t]
\huge
\vspace{-0.4cm}
\begin{center}
\caption{Architecture details of global deblurring discriminator in our proposed framework. }
\label{table:tab_detail_deblurring_discriminator}
\vspace{-0.2cm}
\resizebox{\columnwidth}{!}{
\begin{tabular}{llcc}
\hline\noalign{\smallskip}
Layer $\qquad$& Detail & Output size & Stride\\
\hline\noalign{\smallskip}
Input ($S$ or $D$) $\qquad$ & - $\qquad$& ($H \times W \times 3$) & -\\
Conv \(4 \times 4\) & \textit{IN}+\textit{LeakyReLU} $\qquad$& ($H/2 \times W/2 \times 64$) & 2\\

Conv \(4 \times 4\)& \textit{IN}+\textit{LeakyReLU} $\qquad$& ($H/4 \times W/4 \times 128$) & 2\\
Conv \(4 \times 4\)& \textit{IN}+\textit{LeakyReLU} $\qquad$& ($H/8 \times W/8 \times 256$) & 2\\
Conv \(4 \times 4\)& \textit{IN}+\textit{LeakyReLU} $\qquad$& ($H/16 \times W/16 \times 512$) & 2\\

Conv \(4 \times 4\)& \textit{IN}+\textit{LeakyReLU} $\qquad$& ($H/16 \times W/16 \times 512$) & 1\\
Conv \(4 \times 4\)& \textit{IN}+\textit{Sigmoid} $\qquad$& ($H/16 \times W/16 \times 1$) & 1\\
\hline
\noalign{\smallskip}
\end{tabular}
}
\end{center}
\end{table}

\subsection{Pseudo-blur Synthesizer Training}
The next important step is the learning process of the pseudo-blur synthesizer $\Omega$.
The related modules are reflected in Figure~\ref{fig:fig_mainframework} as red-colored objects.
The reblur module is fed with the deblurred image $D$ or sharp image $S$ to produce the reblurred version $R$.
At the first 50 epochs, we train the reblurrer generator and \textit{global reblur} discriminator using $S$ only to fully learns correct features from the sharp image while treating $B$ as ground truth.
The next 100 epochs are performed with the input data of $D$, which is produced by $\theta_T$ using the dataset of HIDE~\cite{Shen_ICCV2019_Humanaware}.
Finally, in the last 100 epochs, we utilize an additional dataset that placed the human image in the middle region.
In specific, we utilize the pairs of ground truth clean $S$ and blurry $B$ human images from LSP~\cite{Johnson_BMVC2010_LSP} dataset.
For simplicity, we denote this modified dataset as LSPBlur.
Its extraction procedure is discussed in the following discussion.

To produce the desired blurs in LSPBlur, we incorporate a region-separation method that splits both human (foreground) and scene (background) regions.
This operation produces 2 prior binary maps, namely  $\mathcal{M}_u$ and $\mathcal{M}_v$.
The task of $\mathcal{M}_u$ is to cover blurry regions inside and the nearby human body.
This is done as our LSPBlur is defined by human-articulated motion blur (foreground) and scene-motion blur (background).
Simply utilizing any human segmentation algorithm is ineffective as it excludes the blurry region nearby the human body.
The complete procedure of this map extraction is described in the following discussion
Finally, to obtain the scene-blur region map, the reversed version of $\mathcal{M}_v = 1-\mathcal{M}_{u}$ is utilized.

This module is trained using LSGAN~\cite{Mao_ICCV2017_LSGAN}; specifically optimizing the reblurring generator along with the \textit{global}, \textit{human}, and \textit{scene reblur} discriminators.
Both \textit{human} and \textit{scene reblur} discriminators utilize the prior maps $\mathcal{M}_u$ and $\mathcal{M}_v$ to explicitly penalize human and scene regions.
The first 50 and 100 epochs (using $S$ and $D$ from HIDE) only include \textit{global} discriminator.
Full discriminators are employed in the last 100 epochs (using LSPBlur dataset).
Take note that, in this stage, the optimized deblurring network is frozen so that the framework is focused on training the reblurring module.
The details of the pseudo-blur synthesizer training are elaborated in the following passages.

\subsubsection{Training Preliminary}
Training this module is divided into 3 categories: (i) first 50 epochs using $S$ of HIDE dataset, (ii) next 100 epochs using the deblurred result $D$ of HIDE dataset, and (iii) last 100 epochs using the sharp image obtained from our LSPBlur dataset.
The HIDE dataset is chosen as it provides the characteristic of a single sharp image as ground truth and multiple blurry images as output (one-to-many effect). 
Although HIDE dataset~\cite{Shen_ICCV2019_Humanaware} is known for providing human presence, their motion blurs are affected only by the scene blur.
The LSPBlur is utilized to tackle this issue.
\\
\\
\noindent\textbf{Acquiring the LSPBlur}
\begin{figure}[t]
\begin{center}
        \includegraphics[width=0.47\textwidth]{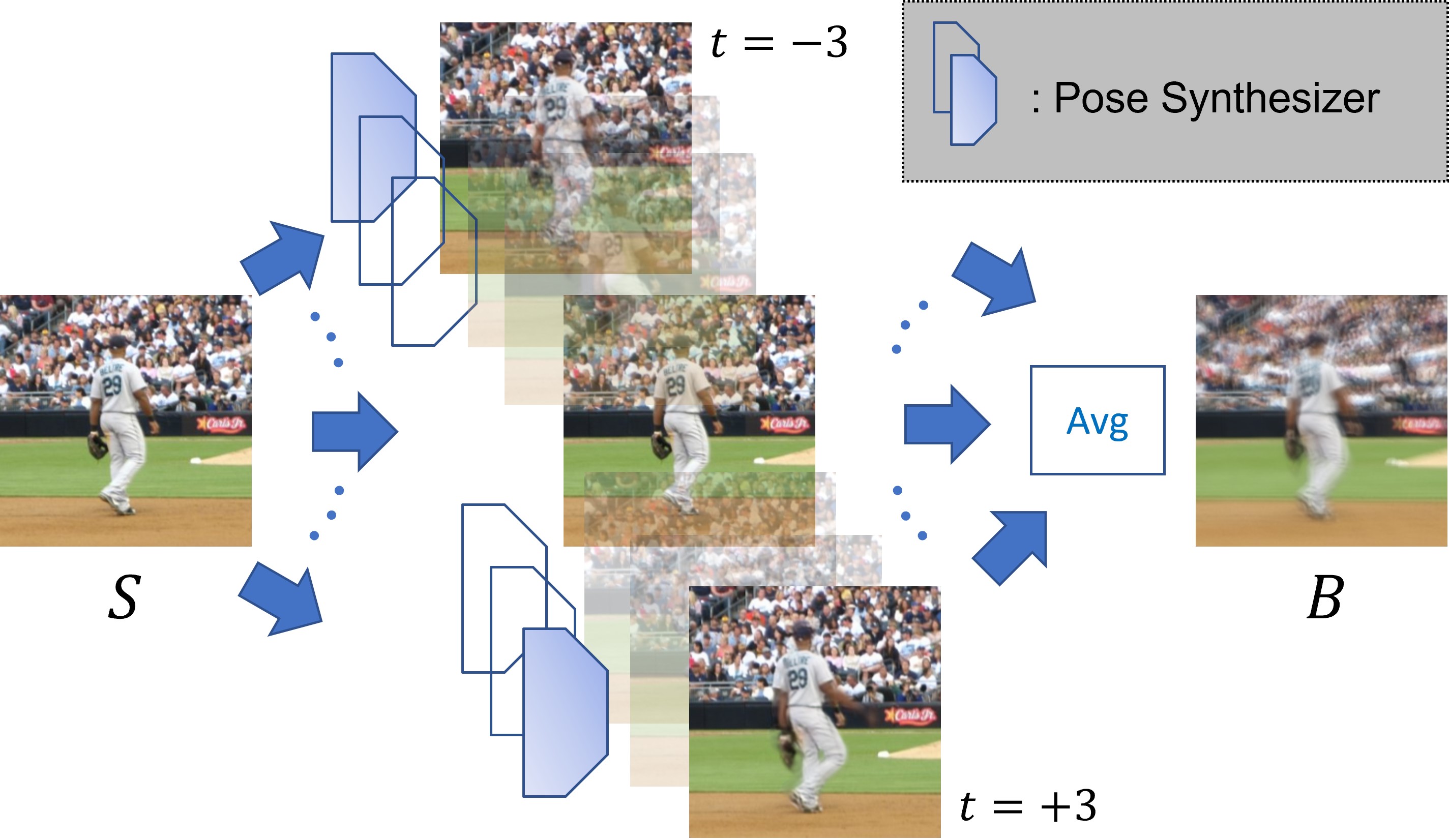}
\end{center}
\vspace{-0.3cm}
\caption{Single sharp image $S$ is utilized to produce multiple images with changed human-pose and translated scene background. These images are averaged together to produce the blurry image $B$.}
\label{fig:fig_synth_pose}
\end{figure}
We realize this drawback and generate a new blurry human dataset that contains human-articulated and scene motions. 
This dataset is collected from the Leeds Sport Dataset (LSP)~\cite{Johnson_BMVC2010_LSP} that contains a human in the middle region of each image. 
Using this image as input, we synthesize a new image with a new human pose and a newly translated scene-background.
The scene translation is generated randomly while the human pose changes are obtained from various Youtube sources.

In detail, from the example of Figure~\ref{fig:fig_synth_pose}, the sharp $S$ image of a man playing baseball is obtained from the LSP dataset.
We then collect the poses of people playing baseball from YouTube video using AlphaPose~\cite{Fang_ICCV2017_Alphapose}, and the pose difference between each video frame is taken as the \textit{change} parameters ($\Delta$).
$\Delta$ values are then utilized to transform the original human pose of the input image $S$ into its new pose printed in the new image at a specific time-stamp $t$.
This procedure is done by employing the \textit{pose-synthesizer} method~\cite{balakrishnan_CVPR18}.
Take note that this work is done for the human part (foreground) while the scene background is translated randomly.
In our experiment, we empirically produce 7 consecutive frames with slightly different body poses and translated backgrounds, as shown in Figure~\ref{fig:fig_synth_pose}.
These images are then averaged together to generate a single blurry output.
This modified dataset (LSPBlur) contains 2,000 pairs of sharp and blurry images.

\subsubsection{Reblurring Module Configuration}
Detailed configurations of our reblurrer module are shown in Table~\ref{table:tab_detail_pseudoblur} while the reblurrer discriminators are displayed in Table~\ref{table:tab_detail_discpseudo}.
Note that the body and scene discriminators receive the input that is masked with the map $\mathcal{M}$ that is extracted using the human prior.
The prior extraction is explained in the next section.
Both deblurring (Tables~\ref{table:tab_detail_deblurring_generator}-\ref{table:tab_detail_deblurring_discriminator}) and reblurring (Tables~\ref{table:tab_detail_pseudoblur}-\ref{table:tab_detail_discpseudo}) modules are processed through the Instance Normalization (\textit{IN}).
In both generators, we utilize 9 Res\_Blocks~\cite{He_CVPR2016} layers.
Each block is constructed by the pattern of Conv$\rightarrow$\textit{IN}$\rightarrow$\textit{ReLU}$\rightarrow$Conv$\rightarrow$\textit{IN} added with initial input.

\begin{table}[t]
\huge
\begin{center}
\caption{Detailed settings of the pseudo-blur synthesizer generator module. The spatial output size is also downsampled and upsampled according to the stride number.}
\vspace{-0.2cm}
\label{table:tab_detail_pseudoblur}
\resizebox{\columnwidth}{!}{
\begin{tabular}{llcc}
\hline\noalign{\smallskip}
Layer $\qquad\qquad$& Detail & Output size & Stride\\
\hline\noalign{\smallskip}
Input ($S$ or $D$) $\qquad$ & - $\qquad$& ($H \times W \times 3$) & -\\
Conv \(7 \times 7\) & \textit{IN}+\textit{ReLU} $\qquad$& ($H \times W \times 64$) & 1\\
Conv \(3 \times 3\)& \textit{IN}+\textit{ReLU} $\qquad$& ($H/2 \times W/2 \times 64$) & 2\\
Conv \(3 \times 3\) & \textit{IN}+\textit{ReLU} $\qquad$& ($H/4 \times W/4 \times 128$) & 2\\
Res\_Blocks\_1-9 \(3 \times 3\) & \textit{IN}+\textit{ReLU} $\qquad$& ($H/4 \times W/4 \times 256$)& 1\\
ConvTrans \(3 \times 3\) & \textit{IN}+\textit{ReLU} $\qquad$& ($H/2 \times W/2 \times 128$) & 2\\
ConvTrans \(3 \times 3\) & \textit{IN}+\textit{ReLU} $\qquad$& ($H \times W \times 64$) & 2\\
Conv \(7 \times 7\) & \textit{Tanh} $\qquad$& ($H \times W \times 3$) & 1\\
\hline
\noalign{\smallskip}
\end{tabular}
}
\end{center}
\end{table}
\begin{table}[t]
\huge
\vspace{-0.4cm}
\begin{center}
\caption{Architecture details of the global, scene, and body reblurring discriminators in our framework.}
\vspace{-0.2cm}
\label{table:tab_detail_discpseudo}
\resizebox{\columnwidth}{!}{
\begin{tabular}{llcc}
\hline\noalign{\smallskip}
Layer $\qquad\qquad$& Detail & Output size & Stride\\
\hline\noalign{\smallskip}
Input ($B$ or $R$) $\qquad$ & - $\qquad$& ($H \times W \times 3$) & -\\
Conv \(4 \times 4\) & \textit{IN}+\textit{LeakyReLU} $\qquad$& ($H/2 \times W/2 \times 64$) & 2\\
Conv \(4 \times 4\)& \textit{IN}+\textit{LeakyReLU} $\qquad$& ($H/4 \times W/4 \times 128$) & 2\\
Conv \(4 \times 4\)& \textit{IN}+\textit{LeakyReLU} $\qquad$& ($H/8 \times W/8 \times 256$) & 2\\
Conv \(4 \times 4\)& \textit{IN}+\textit{LeakyReLU} $\qquad$& ($H/16 \times W/16 \times 512$) & 2\\
Conv \(4 \times 4\)& \textit{IN}+\textit{LeakyReLU} $\qquad$& ($H/16 \times W/16 \times 512$) & 1\\
Conv \(4 \times 4\)& \textit{IN}+\textit{Sigmoid} $\qquad$& ($H/16 \times W/16 \times 1$) & 1\\
\hline
\noalign{\smallskip}
\end{tabular}
}
\end{center}
\vspace{-0.6cm}
\end{table}

\subsubsection{Optimization of Reblurring Module}
\noindent\textbf{Content loss}
The reblurring module is trained to translate $S$ or $D$ into its reblurred version $R$.
The real blurry image $B$ from the dataset is treated as the label. 
In our experiment, the one-to-many effect of HIDE dataset may produce a slight color change between $B$ and $R$.
Therefore, we utilize the Y channel only in the reblurring loss ($L^{reb}_{1}$) after these parameters are converted from RGB to YUV spaces.
The full representation of $L^{reb}_{1}$ is written as follows:
\begin{equation} 
L^{reb}_{1} =  \frac{1}{b}\sum_{c}^{b}\| y(B_c) - y(R_c)\|,
\label{eq:eq_ScoreL1_Reb}
\end{equation}
where $y(\cdot)$ represents the Y channel extraction function.
\\
\\
\noindent\textbf{Human and scene prior extraction}
To fully utilize the adversarial losses, we firstly elaborate the detail of extracting the human and scene prior maps that penalize the input of \textit{body reblur} and \textit{scene reblur} discriminators (refer to Figure~\ref{fig:fig_mainframework}).
The prior is defined as a binary map ($\mathcal{M}$), and its ultimate goal is to find human-motion blur \textit{inside} and \textit{nearby} human body region in an image of our LSPBlur dataset.
In detail, this map is firstly obtained by finding human body keypoints from deblurred image $D$ detected using the sophisticated body predictor module of Kanazawa~\etal~\cite{kanazawa_CVPR18}.
This module extracts the human \textit{body-joint} and \textit{shape} parameters from the input image using the statistical body model~\cite{loper_ACM15}.
Instead of \textit{shape}, we opt to utilize \textit{body-joint} parameter that extracts 14 body keypoints. 
These keypoints are then connected with lines to cover the region \textit{inside} human body.
We denote this map as \textit{body-joint} map.

To obtain the blurry region \textit{nearby} human body, we firstly find the edge difference between $D$ and $R$ using the Sobel filter.
This \textit{difference} map is max-pool-ed to fill the holes and then cropped using the most-top, -right, -bottom, and -left coordinates of the extracted keypoints.
We then combine the \textit{difference} and \textit{body-joint} maps to produce single binary map $\mathcal{M}_{u}$ that fully covers the region \textit{inside} and \textit{nearby} human body.
Its reversed version $\mathcal{M}_{v}$ is utilized to cover the remaining scene region.
\\
\\
\noindent\textbf{Adversarial losses}
Finally, we determine the discriminator losses to distinguish the real and fake reblurred data.
The discriminator losses for the real case of the body, scene, and non-masked (global) images are represented as:
\begin{equation} 
L^{reb}_{\textbf{Real}} = \frac{1}{3b} \sum_{c}^{b} \Pi_{\textbf{Glo}}(B_{c})+\Pi_u(\mathcal{M}_u \odot B_{c})+\Pi_v(\mathcal{M}_v \odot B_{c}).
\label{eq:eq_RealBodySceneGlobal}
\end{equation}
Similarly, we modeled the losses for fake case in the reblurring as follows:
\begin{equation}
L^{reb}_{\textbf{Fake}} = \frac{1}{3b} \sum_{c}^{b} \Pi_{\textbf{Glo}}(R_{c})+\Pi_u(\mathcal{M}_u \odot R_{c})+\Pi_v(\mathcal{M}_v \odot R_{c}).
\label{eq:eq_FakeBodySceneGlobal}
\end{equation}
Finally, we construct them together as generator and discriminator losses through LSGAN~\cite{Mao_ICCV2017_LSGAN} approach as:
\begin{equation} 
L^{reb}_{\textbf{Disc}} = 0.5 \times ( \| L^{reb}_{\textbf{Real}} - 1\|^2 + \| L^{reb}_{\textbf{Fake}}\|^2);
\label{eq:eq_ScoreDisc_Reb}
\end{equation}
\begin{equation} 
L^{reb}_{\textbf{Gen}} = L^{reb}_{1} + 0.5 \times (\| L^{reb}_{\textbf{Fake}} - 1\|^2).
\label{eq:eq_ScoreGen_Reb}
\end{equation}
Note that the LSGAN implementation is only applied up to this stage.
The next meta-learning-based training stages exclude these discriminator modules as both the reblurring and deblurring generator modules are learned.
\\
\\
\begin{figure}[t]
\begin{center}
		\includegraphics[width=1.0\columnwidth]{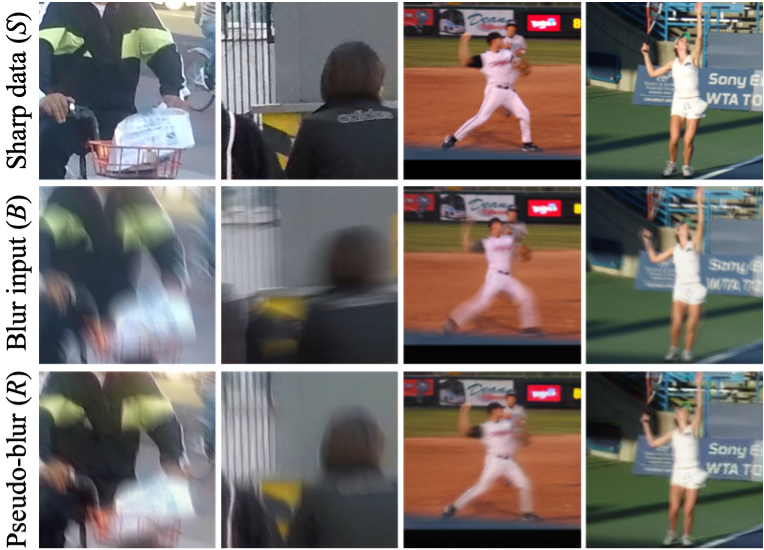}
\end{center}
\vspace{-0.2cm}
\caption{Example of our synthesized blurry results ($R$) or \textit{pseudo-blurred} on the third row using $\Omega_T$ that are \textbf{different} with the blurry input ($B$) in the second row. The first row indicates sharp image input ($S$).}
\label{fig:fig_compsynth}
\end{figure}
\noindent\textbf{Pseudo-blur data examples}
The clear visual representation of our synthesized blur results are demonstrated in Figure~\ref{fig:fig_compsynth}.
Our reblurred results $R$ in the second rows expose different blur patterns than the original blurry input $B$ in the third rows.
The scene blur example is shown in the first two columns from the left.
The articulated motion blur that receives the full human body from the image input is shown in the last two columns from the right.
During training, if a batch is contained with our dataset (LSPBlur), then the global, scene, and human reblur discriminators are utilized.
If the HIDE~\cite{Shen_ICCV2019_Humanaware} dataset is selected, only the global reblur discriminator is employed.
The dataset selection is done randomly at each training iteration.
\begin{algorithm}[t]
\caption{Meta-Transfer Training}
\label{alg:algo_metatransfer}
\textbf{Input}: Pairs of blurry $B$ and sharp images $S$ from data distribution $\mathcal{D}$, reblurring model $\Omega_{T}$, and learning rates $\alpha, \beta$\\
\textbf{Output}: Deblurring model $\theta_{M}$
\begin{algorithmic}[1] 
\STATE Initialize $\theta$ with $\theta_{T}$
\STATE Initialize $\Omega$ with $\Omega_{T}$
\STATE Generate task distribution $p(\mathcal{T})$ from $\mathcal{D}$
\WHILE{not done}
\STATE Sample task batch $\mathcal{T}_{i}^{tr}$ and $\mathcal{T}_{i}^{te}$ from $p(\mathcal{T})$
\FOR{$i$}
\IF{$\mathcal{T}_{i}^{tr}$}
\STATE Data augmentation : \\  $(B^{tr}, S^{tr})\rightarrow (B^{tr}_{*}, S^{tr}_{*})$
\STATE Apply proposed sequence : \\ $B^{tr}_{*} \xrightarrow{\theta} D^{tr}_{in} \xrightarrow{\Omega} R^{tr} \xrightarrow{\theta} D^{tr}_{out}$
\STATE Evaluate task-training loss : \\ $L^{tr}(\theta) = ||S^{tr}_{*}-D^{tr}_{in}||+||y(S^{tr}_{*})-y(D^{tr}_{out})||$ 
\STATE Adapt : $\theta_{i} \leftarrow \theta-\alpha\nabla_{\theta}L^{tr}(\theta)$
\ELSIF{$\mathcal{T}_{i}^{te}$}
\STATE Data augmentation : \\ $(B^{te}, S^{te})\rightarrow (B^{te}_{*}, S^{te}_{*})$
\STATE Apply proposed sequence : \\ $B^{te}_{*} \xrightarrow{\theta_i} D^{te}_{in} \xrightarrow{\Omega} R^{te} \xrightarrow{\theta_i} D^{te}_{out}$
\STATE Evaluate task-testing loss : \\ $L^{te}(\theta_{i}) = ||S^{te}_{*}-D^{te}_{in}||+||y(S^{te}_{*})-y(D^{te}_{out})||$ 
\ENDIF
\ENDFOR
\STATE Update $\theta_{M}$ with respect to average test loss : \\ $\theta_{M} \leftarrow \theta-\beta\nabla_{\theta}\Sigma_{i} L^{te}(\theta_{i}) $
\ENDWHILE
\STATE \textbf{return} meta-transferred $\theta$ : $\theta_{M}$
\end{algorithmic}
\end{algorithm}
\begin{algorithm}[t]
\caption{Meta Testing}
\label{alg:algo_metatester}
\textbf{Input}: Blurry image $B$, meta-transfer trained model $\theta_{M}$, number of gradient updates $n$, and learning rate $\alpha$\\
\textbf{Output}: Deblurred image $D$
\begin{algorithmic}[1] 
\STATE Initialize $\Omega$ with $\Omega_{T}$
\STATE Initialize $\theta$ with $\theta_{M}$
\STATE Initial deblurring : $ B \xrightarrow{\theta} D$
\STATE Augmentation :  $(B, D) \rightarrow B_{*}, D_{*}$
\STATE Apply sequence : $B_{*} \xrightarrow{\theta} D_{in} \xrightarrow{\Omega} R \xrightarrow{\theta} D_{out}$
\FOR{$n$}
\STATE    Evaluate test loss : \\ $L(\theta) = ||D_{*}-D_{in}||+||y(D_{*})-y(D_{out})||$ 
\STATE    Update : $\theta_{k} \leftarrow \theta - \alpha\nabla_{\theta}L(\theta)$
\ENDFOR
\STATE Final deblur using adapted weight : $ B \xrightarrow{\theta_{k}} D$
\end{algorithmic}
\end{algorithm}

\subsection{Meta-transfer-learning for deblurring}
Until the previous step, our method achieves initial deblurring $\theta_T$ and reblurring weights $\Omega_T$, as shown in Figure~\ref{fig:fig_meta_training_scheme}. 
The next objective is to find an optimized deblurring parameter that is suitable for the $\textit{B}\rightarrow\textit{D}\rightarrow\textit{R}\rightarrow\textit{D}$ procedures. 
To achieve it, we apply a meta-transfer-learning operation that seeks to find the initial stable weight $\theta_M$ to be transferred during meta-testing. 
Algorithm~\ref{alg:algo_metatransfer} explains our approach, where lines 6–17 illustrate the inner loop implementation. 
The weight is gradually updated via Gradient Descent optimizer with $\alpha=0.01$ within each task $\mathcal{T}_i$ using $L^{tr}(\theta)$. 
The meta-learner updates the final weight in line 18 using the average of task-test loss ($\Sigma_{i}L^{te}(\theta_i)$) that is optimized using ADAM with $\beta=0.0001$. 
The large learning rate of $\alpha$ is empirically utilized to obtain fast updates on each task. 
$\beta$ is determined with a smaller value to carefully backpropagate through the average of the task-test loss (line 18). 
The $2\times$ downsamplings are applied in lines 8 and 13 aims to simulate a clear blur difference during augmentation.
The $\textit{B}\rightarrow\textit{D}\rightarrow\textit{R}\rightarrow\textit{D}$ is identified at lines 9 and 14 for training ($\mathcal{T}^{tr}_i$) and testing ($\mathcal{T}^{te}_i$) tasks, respectively. 
The loss functions are compared between the deblurring results ($D^{tr}_{in}, D^{tr}_{out}, D^{te}_{in}, D^{te}_{out}$) with the sharp versions ($S^{tr}_{*}, S^{te}_{*},$) from both training and testing tasks, respectively (Lines 10 and 15 of Algorithm~\ref{alg:algo_metatransfer}). 

\begin{figure}[t]
\begin{center}
		\includegraphics[width=1.0\columnwidth]{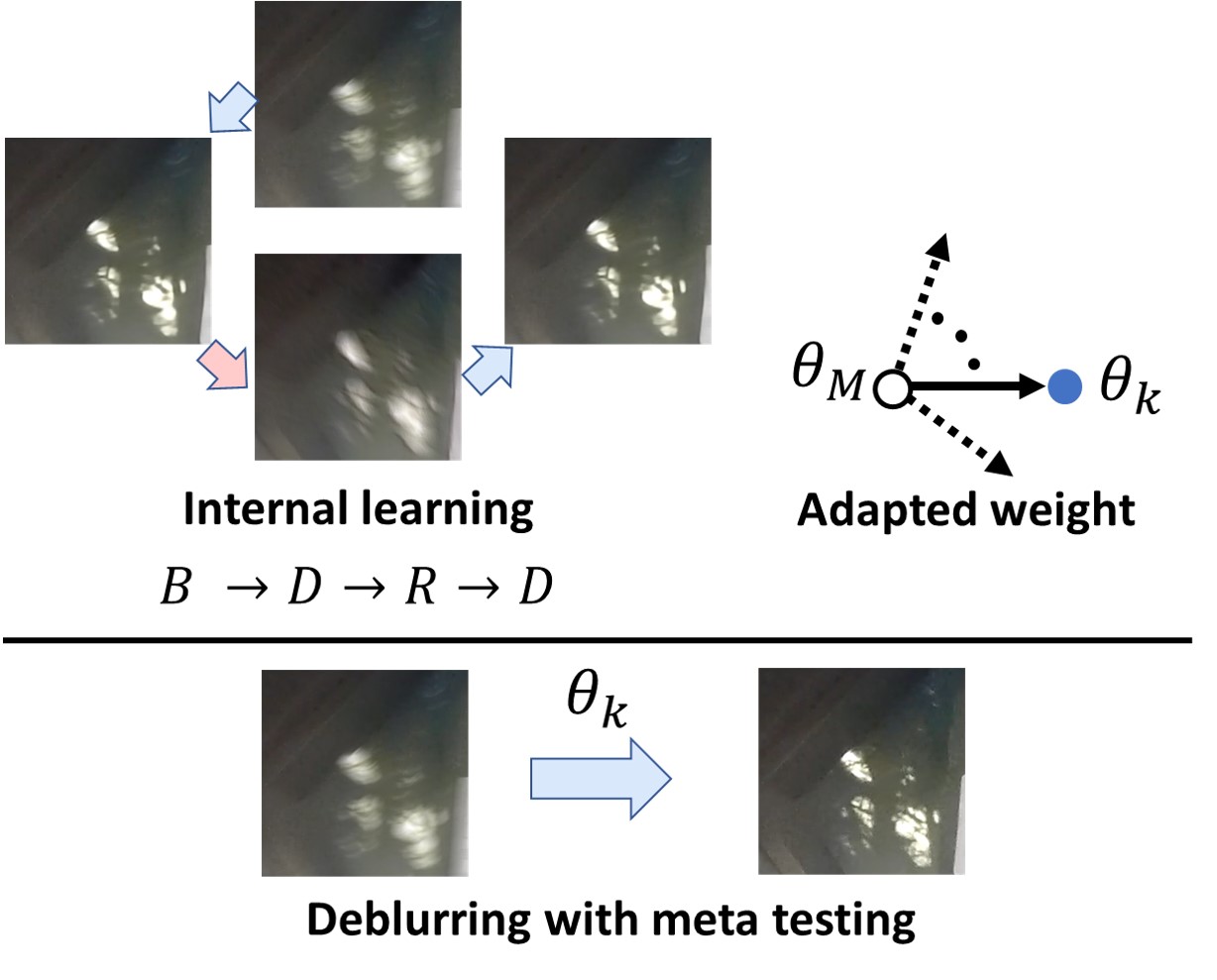}
\end{center}
\vspace{-0.3cm}
\caption{The main scheme of our \textbf{testing} phase. Test-time adaptations ($ \theta_M \rightarrow  \theta_k $) are applied to perform the internal learning using the proposed sequence. Final adapted $\theta_k$ is used for producing final deblurred output.}
\label{fig:fig_meta_testing_scheme}
\end{figure}
\subsection{Meta-Testing for Deblurring}
\label{lab:meta_testing_parag}
The objective of this stage is to perform the testing procedure using the self-adapted weight. 
As illustrated in Figure~\ref{fig:fig_meta_testing_scheme}, the meta-testing procedure of Algorithm~\ref{alg:algo_metatester} transfers the weight of a meta-learned position ($\theta_M$) into its new adapted position ($\theta_k$).
This procedure is done individually in each input data to induce the self-adaptation capability.
Similar to meta-training, our algorithm applies the proposed sequence (Line 5) after the initial deblurring and augmentation.
However, in this test stage, where no sharp label is available, we opt to utilize the initial deblurred result $D_*$ as the supervisor.
As shown in Line 7 (Algorithm~\ref{alg:algo_metatester}), the test-loss $L(\theta)$ is evaluated using the self-extracted data ($D_*, D_{in}, D_{out}$).
The self-adaptation process is performed by gradually updating $\theta$ in the iteration (Line 6-9) using $\alpha = 10^{-2}$ via Gradient Descent optimizer similar to the inner-loop scope of Algorithm~\ref{alg:algo_metatransfer}.
After a certain number of iterations ($n$), the adapted weight $\theta_k$ is utilized for final deblurring (Line 10).

\begin{figure}[t]
\begin{center}
		\includegraphics[width=1.0\columnwidth]{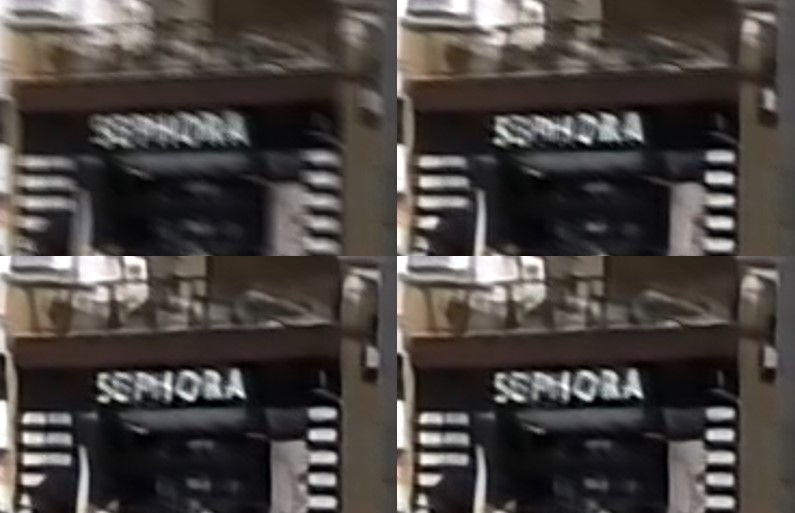}
\end{center}
\vspace{-0.2cm}
\caption{Example of deblurring results from our ablation study. Top-left-to-right: blurry input and Ours-\textbf{F}. Bottom-left-to-right: Ours-\textbf{M}(1) and Ours-\textbf{M}(10).}
\label{fig:fig_ablation}
\end{figure}
\begin{table}[t]
\begin{center}
\caption{
Ablation study results of our approach. The scores are shown to display the pseudo-blur synthesizer's effect in the fine-tuned (-\textbf{F}) and meta-learned (-\textbf{M}) versions of our deblurring.}
\huge
\label{tab:tab_ablation}
\resizebox{\columnwidth}{!}{
\begin{tabular}{|l|c|c|c|c|c|}
\hline
Methods & Ours-\textbf{0} & Ours-\textbf{F} & Ours-\textbf{M}(1) & Ours-\textbf{M}(5) & Ours-\textbf{M}(10)\\
\hline
\hline
SSIM & 0.7588 & 0.8235 & 0.8470 & \color{red}{0.8470} & \color{blue}{0.8472}   \\ 
PSNR & 32.2870 & 33.1490 & 33.4614 & \color{red}{33.4750} & \color{blue}{33.4776}  \\
Time (s) & 0.1446 & 0.1446 & 5.5567 & 27.205 & 54.266  \\
Complex & $O(m)$ & $O(m)$ & $O(mn+m)$ & $O(mn+m)$ & $O(mn+m)$  \\
\hline
\end{tabular}
}
\end{center}
\end{table}

\begin{table*}[t]
\caption{Quantitative comparisons using PSNR and SSIM metrics betweeen deblurring algorithms in HIDE~\cite{Shen_ICCV2019_Humanaware} test set. The blue- and red-colored texts represent the first and second best results.}
\vspace{-0.2cm}
\centering
\small
\begin{tabular}{|l|c|c|c|c|c|c|c|}
\hline
Methods & DebGAN-V1~\cite{Kupyn_CVPR2018} & DebGAN-V2~\cite{Kupyn_ICCV2019}  & DHMP~\cite{ZhangStack_CVPR2019} & HIDE~\cite{Shen_ICCV2019_Humanaware} & SAPH~\cite{Suin_CVPR2020} & Ours-\textbf{F} & Ours-\textbf{M}(1) \\
\hline
\hline
SSIM & 0.7424 & 0.8485  & 0.8376 & \color{blue}{0.9310}  &  \color{red}{0.9300} & 0.8096 & 0.8319   \\ 
PSNR & 29.17 & \color{blue}{33.29}  & 33.12 & 28.89 & 29.98 & 32.96 & \color{red}{33.24} \\
Complex & $O(m)$ & $O(m)$  & $O(m)$ & $O(m)$ & $O(m)$ & $O(m)$ & $O(mn+m)$ \\
\cline{1-6}
\hline
\end{tabular}
\label{tab:tab_hide}
\end{table*}

\begin{table*}[t]
\caption{Quantitative comparisons using PSNR and SSIM metrics betweeen deblurring algorithms in GoPro~\cite{Nah_CVPR2017} test set. The blue- and red-colored texts represent the first and second best results.}
\vspace{-0.2cm}
\centering
\small
\begin{tabular}{|l|c|c|c|c|c|c|c|c|}
\hline
Methods & DebGAN-V1~\cite{Kupyn_CVPR2018} & DebGAN-V2~\cite{Kupyn_ICCV2019}  & DHMP~\cite{ZhangStack_CVPR2019} & HIDE~\cite{Shen_ICCV2019_Humanaware} & SAPH~\cite{Suin_CVPR2020} &  RDeb~\cite{ZhangRealist_CVPR2020} & Ours-\textbf{F} & Ours-\textbf{M}(1) \\
\hline
\hline
SSIM & \color{blue}{0.9580} & 0.9340 & 0.9453 & 0.9400 & \color{red}{0.9530} & 0.9424 & 0.8073 & 0.8253  \\ 
PSNR & 28.70 & 29.55 & 31.20 & 30.26 & 32.02 & 31.10 & \color{red}{32.62} & \color{blue}{32.85}\\
Complex & $O(m)$ & $O(m)$ & $O(m)$ & $O(m)$ & $O(m)$ & $O(m)$ & $O(m)$ & $O(mn+m)$\\
\cline{1-6}
\hline
\end{tabular}
\label{tab:tab_gopro}
\end{table*}

\begin{figure*}[t]
\begin{center}
		\includegraphics[width=0.98\textwidth]{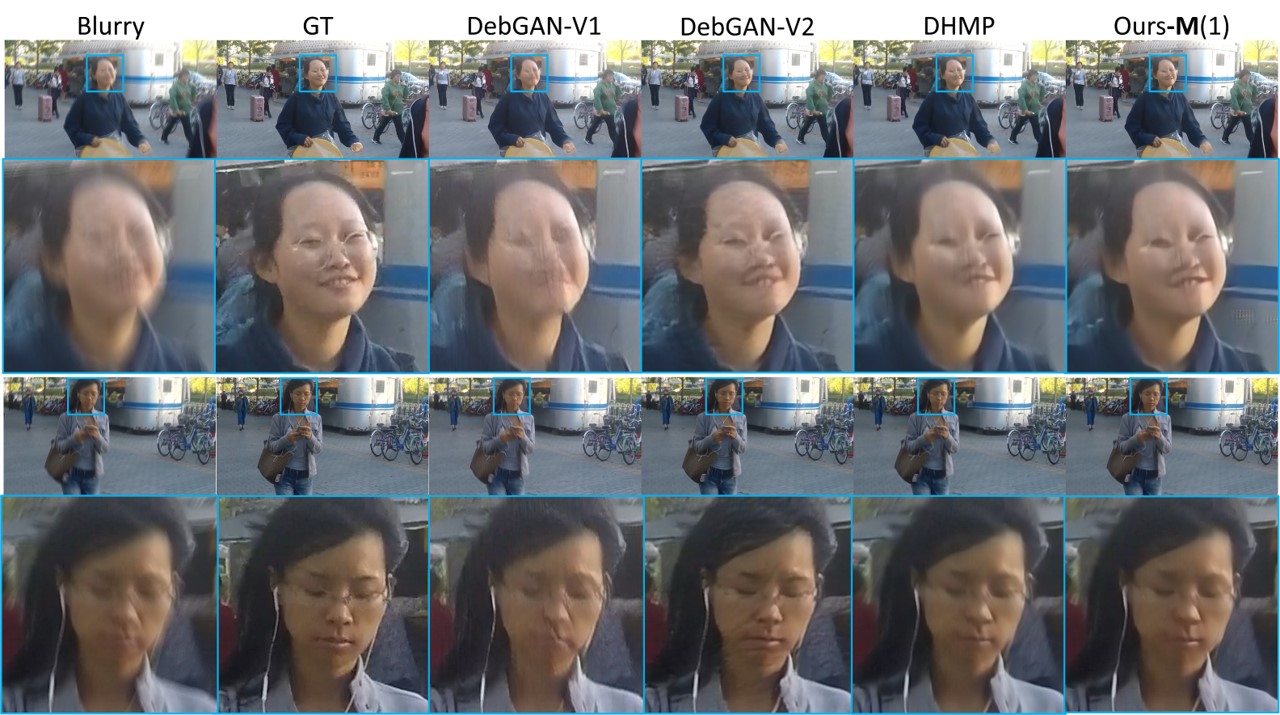}
\end{center}
\vspace{-0.3cm}
\caption{Qualitative results of recent deblurring algorithms compared to our approach using HIDE~\cite{Shen_ICCV2019_Humanaware} test set.}
\label{fig:fig_HIDE_qual}
\end{figure*}
\begin{figure*}[t]
\begin{center}
        \includegraphics[width=0.98\textwidth]{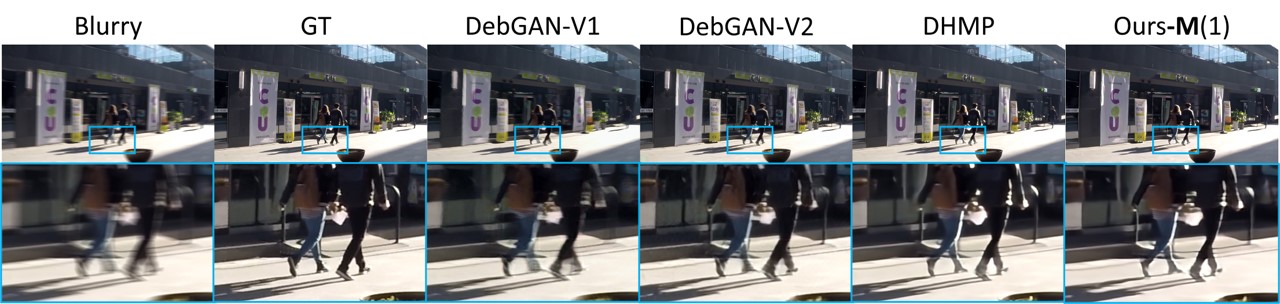}
\end{center}
\vspace{-0.3cm}
\caption{Qualitative results of recent deblurring algorithms compared to our approach using GoPro~\cite{Nah_CVPR2017} test set.}
\label{fig:fig_GOPRO_qual}
\end{figure*}
\begin{figure*}[t]
\begin{center}
        \includegraphics[width=0.98\textwidth]{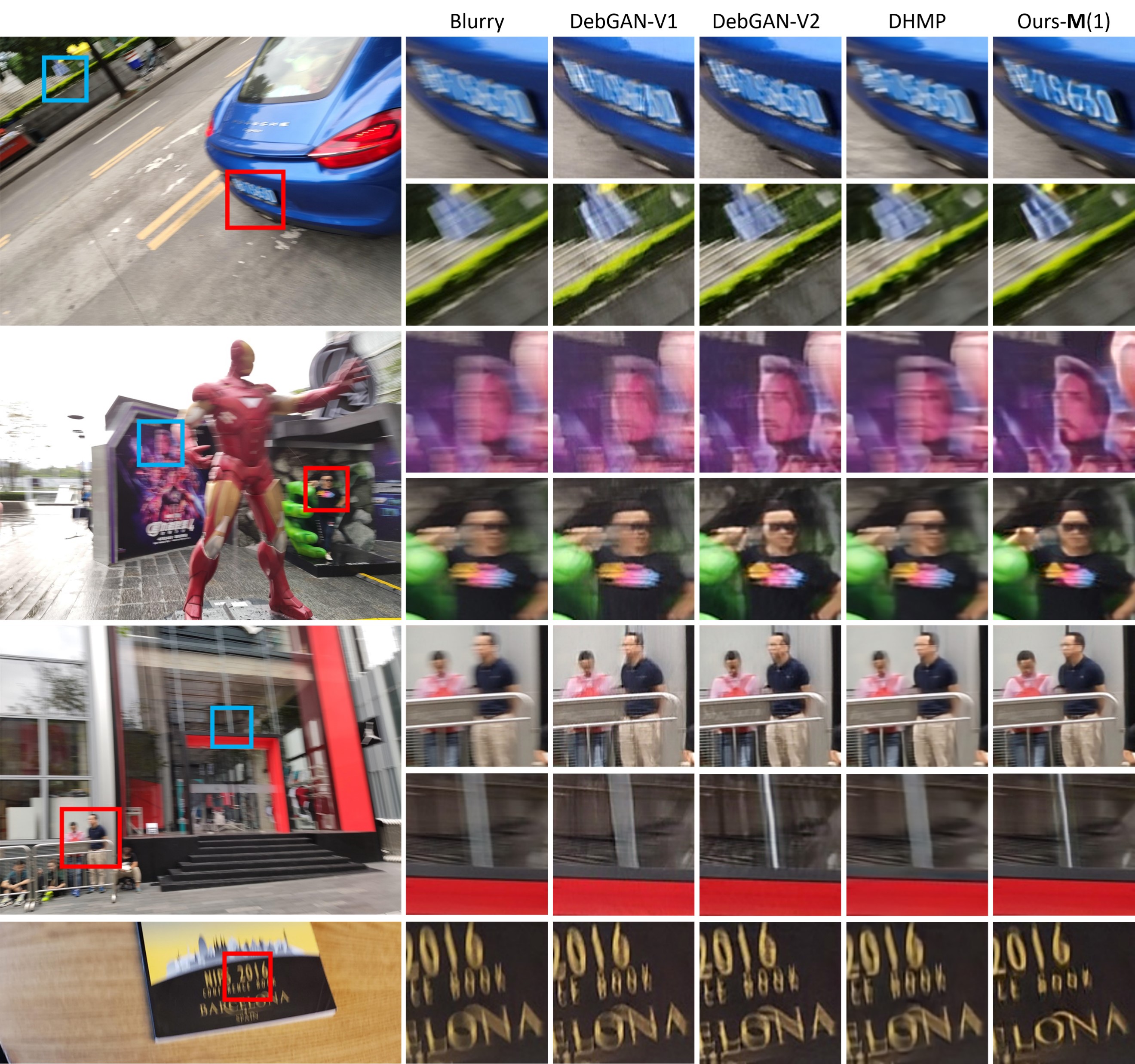}
\end{center}
\vspace{-0.3cm}
\caption{Qualitative results of recent deblurring algorithms compared to our approach using in-the-wild blurry data from RWBI~\cite{ZhangRealist_CVPR2020}.}
\label{fig:fig_RWBI_qual}
\end{figure*}

\section{Experiment}
\subsection{Implementation Details}
\label{lab:implement_detail_parag}
Our implementations are written using TensorFlow and run on a Titan RTX GPU.
The meta-training procedure is fed with $128\times128$ patches and processed with a mini-batch of 8 ($\mathcal{T}^{tr}=4$ and $\mathcal{T}^{te}=4$).
The loss functions of $\left\| * \right\|$ and $\left\| * \right\|^2$ in all equations and algorithms denote standard \textit{absolute} and \textit{mean-squared} errors, respectively.
Our whole training scheme involves 3 particular datasets: (i) GoPro~\cite{Nah_CVPR2017} and  HIDE~\cite{Shen_ICCV2019_Humanaware} for deblurring and (ii) LSPBlur for the reblurring module, respectively.
Total time required to train the networks from $\theta_{0}, \Omega_{0}$ to $\theta_{M}, \Omega_{T}$ (from \textit{initial deblurring training} stage up to \textit{meta-transfer deblurring training} stage refer to Figure~\ref{fig:fig_meta_training_scheme}) is 3 days.

For clarity, we re-explain the parameter details in the following.
In \textit{initial deblurring} training ($\theta_0,\Omega_0\rightarrow\theta_T,\Omega_0$), ADAM with the learning rate of $10^{-4}$ that is utilized.
Same setting with the learning rate of $10^{-4}$ is also applied for the \textit{reblurring} training ($\theta_T, \Omega_0\rightarrow \theta_T,\Omega_T$).
In the \textit{meta-transfer} training scheme for deblurring ($\theta_T, \Omega_T\rightarrow \theta_M,\Omega_T$), the learning rate of $\alpha=10^{-2}$ with Gradient Descent optimizer is utilized in the inner-loop scope, while $\beta=10^{-4}$ and ADAM optimizer are determined in the outer-loop, respectively.
Finally, during \textit{meta-testing}, the self-adaptation of Algorithm~\ref{alg:algo_metatester} utilized $\alpha=10^{-2}$ learned via gradient-descent optimizer.

\subsection{Ablation Study}
\label{sec:subsec_abla_study}
In this work, the key factor that determines the meta-learning performance is the reblurring module.
As the main contribution of this paper is the addition of the reblurring task via pseudo-blur synthesizer module, we provide ablation studies regarding its effect in various conditions, namely: 
\begin{itemize}
    \item Ours-\textbf{0} $\vartriangleright$ No-reblurrer influence in training and testing stages (training/testing: $\textit{B}\rightarrow\textit{D}$).
    \item Ours-\textbf{F} $\vartriangleright$ Reblurrer utilization only in training stage (training: $\textit{B}\rightarrow\textit{D}\rightarrow\textit{R}\rightarrow\textit{D}$; testing: $\textit{B}\rightarrow\textit{D}$).
    \item Ours-\textbf{M} $\vartriangleright$ Full utilization of reblurrer in training and testing stages (training/testing: $\textit{B}\rightarrow\textit{D}\rightarrow\textit{R}\rightarrow\textit{D}$).
\end{itemize}
Ours-\textbf{F} utilizes a naive-learning strategy, where trained reblurrer $\Omega_T$ is utilized to further fine-tunes $\theta_T$ (along with \textit{global deblur} discriminator) by providing the additional augmented blurred data $R$ during training.
We demonstrate this ablation study on a recent HIDE dataset~\cite{Shen_ICCV2019_Humanaware} as it focuses on human and scene motion deblurring cases. 
We utilize the \textit{long-shot} test-cases of~\cite{Shen_ICCV2019_Humanaware} where the scores of each condition is shown in Table~\ref{tab:tab_ablation}.
Number of adaptation $n$ (scripted in Line 6 of Algorithm~\ref{alg:algo_metatester}) is placed next to each Ours-\textbf{M} method. 
In our results, increasing the number of adaptations ($n$) improves the performance and indicates that Ours-\textbf{M} setting succeeds in performing self-adaptation using the meta-transferred weight.
The proposed work that utilizes self-adaptation, Ours-\textbf{M}, performs superior compared to the naive fine-tuned (Ours-\textbf{F}) and the classic versions (Ours-\textbf{0}) where the qualitative evidences are shown in Figure~\ref{fig:fig_ablation}.

Moreover, to clearly display the performance, we also included the time and complexity scores as shown in Table~\ref{tab:tab_ablation}.
The complexity formula is defined by big $O(.)$ notation with the parameters of $m$, which denotes the number of tested data, and $n$, the number of adaptations.
Without any self-adaptation (Ours-\textbf{0} and Ours-\textbf{F}), the execution is only affected by number of data ($m$).
Each test data ($m=1$) requires the computational footprint of 0.1446 seconds in a TITAN RTX GPU for processing a $1280\times720$ image.
On the case of Ours-\textbf{M}, the computational time required in performing single-adaptation (Ours-\textbf{M}(1) or $n=1$) is 5.4121 seconds.
Thus, the total time required for performing the \textit{adaptations} and \textit{final meta-testing} ($O(mn+m)$) of Algorithm~\ref{alg:algo_metatester} is 5.5567 seconds.
As the quantitative scores of Ours-\textbf{M}(1), (5), and (10) are marginal (shown in Table~\ref{tab:tab_ablation}), we utilize $n=1$ in the further experiment to avoid large time-consumption.
In the next discussions, we elaborate our experiment using the common benchmark dataset (train-available dataset) and recent real-world motion-blurred dataset.

\subsection{Test Using Train-Available Dataset}
In this stage, we demonstrate the performance of our approach along with other state-of-the-art deblurring methods of DebGAN-V1~\cite{Kupyn_CVPR2018}, DebGAN-V2~\cite{Kupyn_ICCV2019}, DHMP~\cite{ZhangStack_CVPR2019}, HIDE~\cite{Shen_ICCV2019_Humanaware}, SAPH~\cite{Suin_CVPR2020}, and RDeb~\cite{ZhangRealist_CVPR2020}.
We utilize the test set of recent blurry human (HIDE)~\cite{Shen_ICCV2019_Humanaware}, and general (GoPro)~\cite{Nah_CVPR2017} dataset for measuring the quantitative performance.
The quantitative scores of this experiment are reflected in Table~\ref{tab:tab_hide} for the HIDE case and Table~\ref{tab:tab_gopro} for the GoPro case. 
Blue and red colors annotate the top 1 and 2 achievers. 
The qualitative results are visualized in Figures~\ref{fig:fig_HIDE_qual} and ~\ref{fig:fig_GOPRO_qual} for HIDE and GoPro cases, respectively.

In the quantitative measurement, we include: Ours-\textbf{F} and Ours-\textbf{M}(1), as both approaches have the assistance of the blurry synthesizer. 
In the HIDE test case, as seen in Table~\ref{tab:tab_hide}, Ours-\textbf{M} achieves a high PSNR result compared with the recent algorithms. 
The score of Ours-\textbf{M} is faithful enough as it achieves similar PSNR compared with the recent deep architecture method~\cite{ZhangStack_CVPR2019}. 
Our high quantitative scores on the HIDE case are supported by the results in Figure~\ref{fig:fig_HIDE_qual}.
From the observation on the electronic screen, our qualitative results in Figure~\ref{fig:fig_HIDE_qual} are close to DebGAN-V2~\cite{Kupyn_ICCV2019} in a positive manner. 
DebGAN-V2~\cite{Kupyn_ICCV2019} provides clear restored edge output; however, it suffers from an artifact that is seen in the homogeneous region (\eg, face in Figure~\ref{fig:fig_HIDE_qual}). 
Since DebGAN-V2~\cite{Kupyn_ICCV2019} is optimized via dual discriminators, we believe this artifact is a product of the synthesizing procedure rather than the restoration.
Our method that is also coupled with multiple discriminators also produces a similar issue; yet, it still preserves realistic output in these regions.
This benefit is obtained as our method initially learns the internal features of the input image during the test stage (Line 8 of Algorithm~\ref{alg:algo_metatester}).

In the GoPro~\cite{Nah_CVPR2017} case, we show that Ours-\textbf{M} achieves the best performance in terms of PSNR score (Table~\ref{tab:tab_gopro}).
Although both datasets contain humans, the GoPro~\cite{Nah_CVPR2017} dataset is closely related to the scene and human-articulated motion blur scenario.
This characteristic is obtained because the blur in the GoPro dataset is extracted from the sequential video frames that automatically capture the natural scene and human motion blurs.
As shown in Figure~\ref{fig:fig_GOPRO_qual}, all deblurring results in the GoPro case are visually similar.
While the proposed work is robust in restoring the blurry input, our deblurring approach came with two limitations, namely: realistic but unreliable structures and longer execution times.

In the first case, our deblurring module is prone to extract structures that are unmatched to the ground truth sharp image, which cost the SSIM scores to be low in all quantitative measurements (Tables~\ref{tab:tab_ablation}-\ref{tab:tab_gopro}).
We believe this anomaly is caused by the GAN method that basically synthesizes the deblurring output rather than restoring it.
Our GAN-based approach is crafted with multiple, precisely 4, discriminators that contribute fully to each initial deblurring and reblurring training.
Nonetheless, our deblurring approach still produces realistic results with consistent color formation, as expressed by the competitive yet high PSNR scores.
As argued in the previous discussion, this achievement is obtained through the internal learning procedure, which includes the pseudo-blur synthesizer role through the reblurring.

In the second case, our method requires longer runtime as it is designed to perform test-time improvements through several $n$ adaptations (our complexity is $O(mn+m)$ as described in the ablation study subsection).
The other methods have their own unique architectures that are designed to solve dynamic scenes. 
Our deblurring generator module, on the other hand, is designed equally to DebGAN-V1~\cite{Kupyn_CVPR2018} which guarantees that our major contributions are located on the reblurrer and the algorithms themselves. 
Other methods that only rely on single-forward pass surely surpass our runtime since their execution is only affected by the number of test-data $m$.
By utilizing single batch in test-time, the iterations of other methods only reach the complexity of $O(m)$ in Tables~\ref{tab:tab_hide}-\ref{tab:tab_gopro}.
However, we believe our case with $n=1$ iteration (Ours-\textbf{M}(1)) is still tolerable as our ultimate goal is to invoke the self-adaptive capability of the deblurring module in solving \textit{unknown} data.

As seen in Tables~\ref{tab:tab_hide}-\ref{tab:tab_gopro}, the recent deblurring methods are already adjusted to this scenario as HIDE~\cite{Shen_ICCV2019_Humanaware} and GoPro~\cite{Nah_CVPR2017} test sets are extracted in a similar way to their training sets.
Certainly, the compared state-of-the-art works are able to solve them with marginal scores.
The recent deblurring methods crafted with sophisticated deep-learning functions are already robust to these benchmarks.
To fully witness the self-adaptive capability of our approach, other unknown test cases not relevant to the training set (GoPro~\cite{Nah_CVPR2017} and HIDE~\cite{Shen_ICCV2019_Humanaware}) are demanded.
In the next experiment, we perform an in-the-wild deblurring test using Ours-\textbf{M}(1) to tackle this issue.

\subsection{Test Using In-the-Wild Dataset}
Our method's superiority is reflected in the scenario of real-world (in-the-wild) blurry data restoration.
To demonstrate it, we utilize the recent Real-World Blurry Images (\textbf{RWBI}) dataset released by Zhang~\etal~\cite{ZhangRealist_CVPR2020}, where no training nor sharp ground truth data are available.
This dataset is captured using various devices, namely: iPhone XS, Samsung S9 Plus, Huawei P30 Pro, and GoPro Hero 5 Black.
The previous deep-learning-based deblurring methods mostly rely on the provided training set.
This practice seems limited as the real-world blurry case can differ in terms of motion blur patterns or scales or color distribution.

As shown in Figure~\ref{fig:fig_RWBI_qual}, other deblurring methods suffer from several issues.
To be precise, the method of DebGAN-V1~\cite{Kupyn_CVPR2018} fails to preserve consistent color information where the deblurred version tends to be brighter than its blurry input.
Large-scale blur pattern cases are also not solved using this method.
Its improved version, DebGAN-V2~\cite{Kupyn_ICCV2019}, solves the consistency but still lacks to restore large motion blur pattern.
Another recent work by DHMP~\cite{ZhangStack_CVPR2019} seems to suffer in restoring regions with strong and weak edges.
This issue is shown by large and small blur patterns that are failed to be restored by the recent methods.

This phenomenon shows that Ours-\textbf{M} is more reliable and adaptive than other methods ~\cite{Kupyn_CVPR2018,Kupyn_ICCV2019,ZhangStack_CVPR2019} that still utilize traditional way ($\textit{B}\rightarrow\textit{D}$).
As shown in Figure~\ref{fig:fig_RWBI_qual}, our approach is able to restore various cases, including text, human, and other object's motion, with consistent results.
By this exploration, we believe that the proposed sequence can be utilized in any deblurring scenario, although the training data is limited.
More convincing results are provided in our supplementary video.
Readers are encouraged to check all results on an electronic screen.

\section{Conclusion}
In this work, we present an unorthodox approach of deblurring under $\textit{B}\rightarrow\textit{D}\rightarrow\textit{R}\rightarrow\textit{D}$ strategy, which includes in-the-fly reblurring operation during training and testing stages. 
To achieve it, we supply our network with a pseudo-blur synthesizer module. 
The synthesizer acts as a blurry data augmenter, which helps improve the deblurring network’s performance. 
To form a reliable blurry synthesizer, we opt to utilize hand-crafted prior extracted from the human statistical model. 
Its objective is to let the network learns to produce human-articulated and scene motion blurs simultaneously. 
This is achieved in the image spatial domain where blurry regions of scene and human are distinguished by the prior. 

Finally, we show that by employing the blur synthesizer, the deblurring module learns new blur information, which subsequently improves the performance.
This benefit is clearly shown in our experiments, especially in restoring the real-world blurry data.
We believe this finding is gainful in deblurring studies where only limited training data is available.
Furthermore, our method emboldens future works to shift from traditional way ($\textit{B}\rightarrow\textit{D}$) to the more-adaptive proposed sequence ($\textit{B}\rightarrow\textit{D}\rightarrow\textit{R}\rightarrow\textit{D}$) for deblurring.
We leave the study of the best network architecture that fits our sequence as part of future contributions.

\bibliographystyle{IEEEtran}
\bibliography{IEEEabrv, bib_pseudo_blur}

\newpage
\vspace*{-12cm}
\begin{IEEEbiography}[{\includegraphics[width=1in,height=1.25in,clip,keepaspectratio]{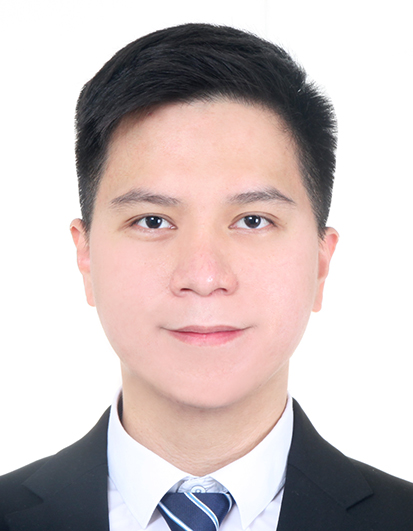}}]{Jonathan Samuel Lumentut} received his B.CompSc and M.S. degree in computer science from Bina Nusantara University, Indonesia, in 2013 and 2014, respectively. He was a visiting student with the Center of Visual Computing, University of California San Diego (UCSD). He received his Ph.D. degree in the Electrical and Computer Engineering from Inha University, Korea, in 2021. His research interests include computer vision, computational photography, human body reconstruction, and image processing. He is a member of the IEEE.
\end{IEEEbiography}

\vspace*{-12cm}
\begin{IEEEbiography}[{\includegraphics[width=1in,height=1.25in,clip,keepaspectratio]{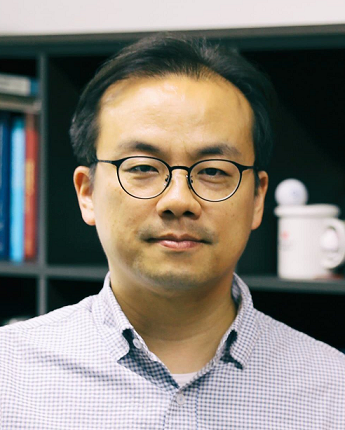}}]{In Kyu Park} (S'96-M'01-SM'14) received the B.S., M.S., and Ph.D. degrees from Seoul National University in 1995, 1997, and 2001, respectively, all in electrical engineering and computer science. From September 2001 to March 2004, he was a Member of Technical Staff at Samsung Advanced Institute of Technology. Since March 2004, he has been with the School of Information and Communication Engineering, Inha University, where he is a full professor. From January 2007 to February 2008, he was an exchange scholar at Mitsubishi Electric Research Laboratories. From September 2014 to August 2015, he was a visiting associate professor at MIT Media Lab. From July 2018 to June 2019, he was a visiting scholar at the Center for Visual Computing in University of California, San Diego. Dr. Park's research interests include the joint area of computer vision and graphics, including 3D shape reconstruction from multiple views, image-based rendering, computational photography, deep learning, and GPGPU for image processing and computer vision. He is a senior member of IEEE and a member of ACM.
\end{IEEEbiography}

\EOD

\end{document}